\RequirePackage{fix-cm}
\documentclass[twocolumn]{svjour3}          
\smartqed  
\usepackage{graphicx}
\usepackage{subfigure}
\usepackage{amsmath,amssymb,amsfonts}

\newcommand*{\affaddr}[1]{#1}
\newcommand*{\affmark}[1][*]{\textsuperscript{#1}}
\usepackage[caption=false]{subfig}
\usepackage[usenames]{color}
\usepackage[linesnumbered,ruled,vlined]{algorithm2e}
\definecolor{gray}{RGB}{192,192,192}
\begin{document}

\title{Custom Distribution for Sampling-Based Motion Planning
}


\author{%
		Gabriel O. Flores-Aquino \protect\affmark[1]         \and
        J. Irving Vasquez-Gomez\affmark[2,]\affmark[3]        \and
        Octavio Gutierrez-Frias\affmark[1] 
}%

\authorrunning{Short form of author list} 
\institute{Gabriel O. Flores-Aquino  \at
              \email{gfloresa0500@alumno.ipn.mx}           
           \and
           J. Irving Vasquez-Gomez \at
              \email{jvasquezg@ipn.mx}
           \and
           O. Octavio Gutierrez-Frias \at
              \email{ogutierrezf@ipn.mx} 
           \and           
           \affaddr{\affmark[1]Instituto Polit\'ecnico Nacional (IPN), Secci\'on de Estudios de Posgrado e Investigaci\'on de la Unidad Profesional Interdisciplinaria en Ingenier\'ia y Tecnolog\'ias Avanzadas (UPIITA), Ciudad de M\'exico, M\'exico}\\
           \affaddr{\affmark[2] Instituto Polit\'ecnico Nacional (IPN), Centro de Inovaci\'on y Desarrollo Tecnol\'ogico en C\'omputo (CIDETEC), Ciudad de M\'exico, M\'exico}\\
           \affaddr{\affmark[3]Consejo Nacional de Ciencia y Tecnolog\'ia (CONACYT) Ciudad de M\'exico, M\'exico}
}

\date{Received: date / Accepted: date}

\maketitle

\begin{abstract}
Sampling-based motion planning algorithms are widely used in robotics because they are very effective in high-dimensional spaces. However, the success rate and quality of the solutions are determined by an adequate selection of their parameters such as the distance between states, the local planner, and the sampling distribution. For robots with large configuration spaces or dynamic restrictions, selecting these parameters is a challenging task. This paper proposes a method for improving the performance to a set of the most popular sampling-based algorithms, the Rapidly-exploring Random Trees (RRTs) by adjusting the sampling method. The idea is to replace the uniform probability density function (U-PDF) with a custom distribution (C-PDF) learned from previously successful queries in similar tasks. With a few samples, our method builds a custom distribution that allows the RRT to grow to promising states that will lead to a solution. We tested our method in several autonomous driving tasks such as parking maneuvers, obstacle clearance and under narrow passages scenarios. The results show that the proposed method outperforms the original RRT and several improved versions in terms of success rate, tree density and computation time. In addition, the proposed method requires a relatively small set of examples, unlike current deep learning techniques that require a vast amount of examples.  
\end{abstract}
\section{Introduction}
\label{intro}

\begin{figure}[tb]
\centering
\includegraphics[width=\linewidth]{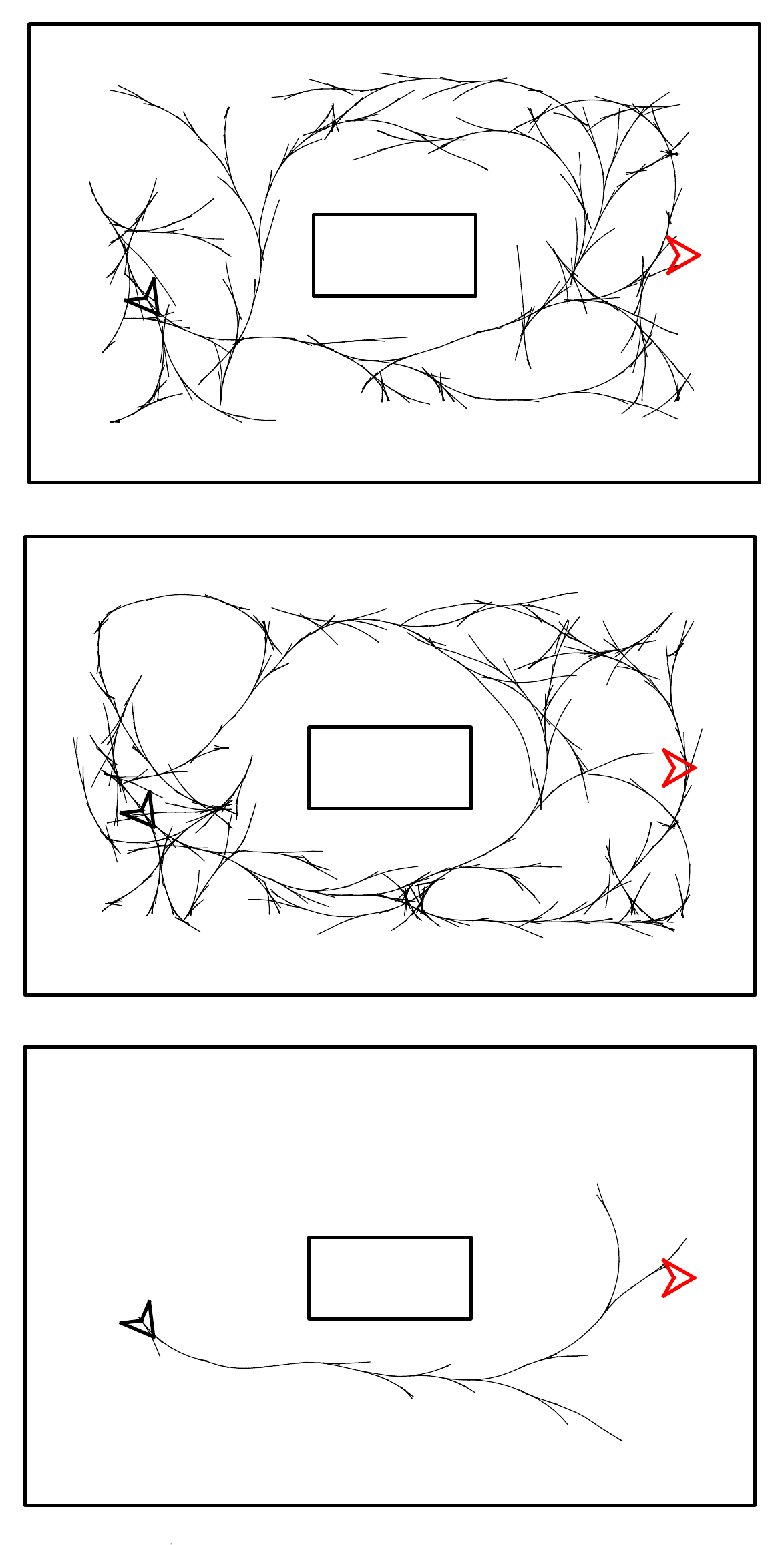}
\label{uTreeTask2}
\caption{\textcolor{black}{Resulting trees for the task of avoiding a static obstacle. From top to bottom, we show the tree that resolves the query, in 3415 vertices with samples from a U-PDF. Follow by the tree that resolves the query, in 4732 vertices with samples from a goal bias distribution (GB-PDF).  In the bottom row, we found a solution for a tree in only 241 vertices with samples from C-PDF. Black triangle indicates starting configuration. The red triangle indicates goal configuration.}}
\label{examplesTreesA}
\end{figure}          

The motion planning problem can be defined as the task of determining the set of inputs for an open-loop control that drives the robot from an initial state to a goal state in a collision-free state space while the kinematic and differential constraints are satisfied \cite{b1}. Motion planning has been investigated from different perspectives \cite{b14} using techniques such as potential fields, neural networks \cite{b29} or genetic algorithms \cite{b30} to name a few. In particular, we are interested in motion planning for autonomous vehicles where the problem has been addressed from formal approaches like control theory \cite{b27} to machine learning. For a complete review, the reader is referred to  \cite{b26}. 

In the last decades, the attention has been focused on sampling-based motion planners (SBMP) due to their simplicity and good performance even in high dimensional spaces \cite{b24}. For sampling-based methods, there are two main approaches: the single-query models and multi-query models. For example, the rapidly exploring random tree (RRT)\cite{b5} and the probabilistic roadmap (PRM)\cite{b4} respectively. Based on them, a large variety of ideas have emerged to cover the weak aspects in order to improve the quality of the solutions or to reduce the computational time. A good example is the RRT*\cite{b12} variation where the solution converges to the optimum and more recently the \textcolor{black}{Stable Sparse RRT (SST)}\cite{b10} method that produces sparse trees or the \textcolor{black}{Fast Marching Tree (FMT*)}\cite{b28} for complex motion planning problems. Other works have been focused on tuning the algorithm parameters; for example, recent methods use reinforcement learning to establish the distance metric \cite{b6}, the local planner strategy \cite{b13} or the probability distribution \cite{b7}. Unfortunately, these approaches require a very large number of examples to infer knowledge, because they are based on deep learning. Nevertheless, we share the enthusiasm of exploring the cross-field between motion planning and machine learning.  \textcolor{black}{Some examples in this field are: in \cite{b29}, the method uses a neural planner, in \cite{b25} the planning is carried out in a learned representation that reduces the state space and in \cite{b32}, a convolutional neuronal network (CNN) is used to generate a non-uniform sampling.} 

This paper studies the way by which the planner samples the state space. In this sense, typically a uniform sampling distribution (U-PDF) is the first option because it produces a graph or roadmap that spans evenly over the free state space. However, in some cases it is not the best choice. For example, in planners such as the PRM; a uniform distribution has poor performance in environments where there are narrow passages because these regions are poorly sampled. To solve the aforementioned problem, some techniques oversample such passages, e.g. the bridge test \cite{b9} or the medial axis method\cite{b34}, which allows refining the connectivity in narrow passages or the Gaussian sampling strategy \cite{b15}, \cite{b8}, whose objective is to obtain a model of the free state spaces, including these difficult passages; other approaches focus on extracting local characteristics from the workspace for biasing the sampling \cite{b17}, \cite{b18}, \cite{b19}. These hybrid sampling strategies are appropriate for planners with a multi-query model like PRM because they maintain a connected roadmap, and this offers the possibility of reusing the roadmap for different queries. But for a single-query model it may be impractical because each query needs to create a new graph hence covering the complete work-space might be unnecessary, while it is more important to achieve a high convergence rate and low computational cost. For this reasons, it is very important to explore new ways to efficiently sample the state space, specially when a small dataset of examples is available. \textcolor{black}{Some works combine techniques from different areas, e.g. \cite{b22}, where the authors combine the advantages of SBMP with graph-based sampling techniques.}  

For a single-query model like the RRT or any tree-based algorithm, the traditional approach consists of sampling the space randomly with a U-PDF and growing the tree iteratively until the goal state is reached. One way to improve the performance is to find a method for guiding the tree growth. For example, in \cite{b16} and \cite{b20}, the tree is guided by selecting the option of expansion with maximal expected utility but, this type of method modifies the structure of the algorithm, which can be hard to implement. Other works, such as \cite{b21}, reuse the found paths by developing a framework to improve the performance with respect to the computing time.

\textcolor{black}{
In this way, recent works use the knowledge from previous experience to improve performance. This emerging field, denominated experience based-planning \cite{b31}, takes vantage on the suppose that the workspace is similar in multiples tasks. Therefore, previous paths can be useful for futures tasks. For
example, in \cite{b23} and \cite{b33} the previous knowledge is used to reduce the query resolution time.}

The proposed method uses the knowledge from previous queries to improve the sampling function. Our hypothesis is that for each environment there is a more appropriate sampling function different from the uniform distribution. It focuses on tuning the probability density function used in the sampling method so that it builds sparser trees and at the same time it increases the success rate with relatively few samples. See Figure \ref{examplesTreesA} as example. In other words, we are interested in guiding the tree by only replacing the sampling function without making further modifications to the algorithm. We have compared our method versus the classic version of RRT with a uniform distribution and versus an RRT version with the distribution skewed by the target state (RRT-goal-bias) \textcolor{black}{in five tasks, task 1 is a low speed obstacle avoidance, tasks 2 and 3 are parking manoeuvres, task 4 is a more complex environment with narrow passages and task 5 is a high speed obstacle avoidance}. The experiments show that our method generates directed trees that are especially useful in environments where the robot has a similar goal but its initial state varies. \textcolor{black}{Direct trees imply a lower computing time; for robotic tasks where we need a solution path almost immediately, reducing the planning time while maintaining or improving the success rate for the algorithm is a critical problem for autonomous navigation. For this reason, the proposed experiments are useful tasks for autonomous driving cars, like autonomous parking and evasive maneuvers at low and high speeds}. Our method minimizes the effects of a not well-adapted sampling domain and has significant improvement over classic RRT. In addition, our results provide evidence that the method can be applicable across other planners or robotic platforms.

The rest of the paper is organized as follows. In section \ref{theoricalBackground}, we present the generality of RRT algorithm and its parameters for the three proposed environments. In section \ref{cpdfMethod}, we make the description of the proposed method. In section \ref{evaluation}, we present the numerical results about our custom distribution compared against the classic RRT and goal-bias version. Finally, in section \ref{conclusion}, we present the conclusions and future work.    

\section{Theoretical background}
\label{theoricalBackground}

This section briefly describes the RRT algorithm and its parameters such as the robot model, the distance metric and the environment for the proposed tasks.  

\subsection{Robot model}

\textcolor{black}{For the presented experiments, we define the robot's State Transition Equation (STE) with a kinematic model (in sub-section \ref{subsec_kinematiModel}) for tasks with low-speed profile and as a dynamic model (in sub-section \ref{subsec_dynamicModel}) for tasks that involve a high-speed profile.}

\subsubsection{Kinematic model}
\label{subsec_kinematiModel}
\begin{figure}[tb]
\begin{center}
\includegraphics[width = \linewidth]{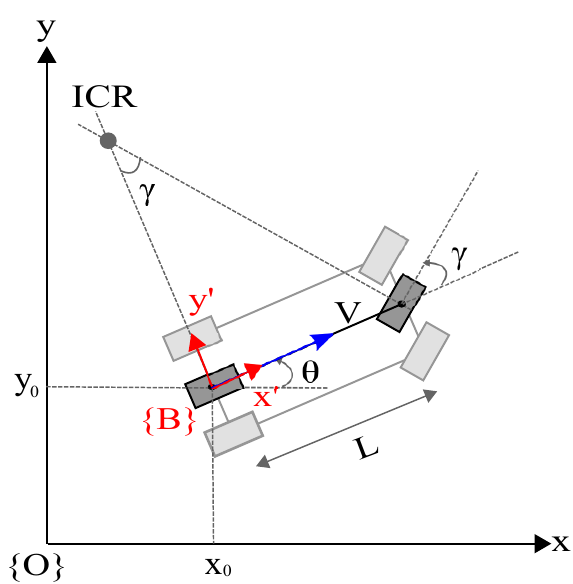}
\end{center}
\caption{Kinematic model of a car. The world frame $\{O, x, y\}$ in color black. The body frame $\{B,x', y'\}$. ICR denotes the Instantaneous Center of Rotation and $L$ the distance between wheels axles.}
\label{kinematicModel}
\end{figure}

\textcolor{black}{For the kinematic model, the STE is defined by a non-linear ordinary differential equation presented in column vector form in equation (\ref{ete}). The reader can be deduced this equation from the diagram in Figure \ref{kinematicModel}.}

\begin{equation}
f(x, u) = \begin{bmatrix}
\dot{\mathrm{x}} \\
\\
\dot{\mathrm{y}} \\
\\
\dot{\theta} \\
\end{bmatrix} = \begin{bmatrix}
V \cos (\theta) \\
\\
V \sin (\theta)\\
\\
V \frac{ \tan(\gamma)}{L} \\ 
\end{bmatrix}
\label{ete}
\end{equation}

\textcolor{black}{
Let $\mathcal{X}$ denoted the state space such that $x \in \mathcal{X}$, where $x$ is the set of coordinates in the inertial frame. Then, $x_o$ and $y_o$ define position on the plane and $\theta_{o}$ defines the car orientation. This model is suitable for tasks that involve low velocity of translation where it is not necessary considerate a tire model. From equation (\ref{ete}) the variable $u$ represents the robot input such that $u \in \mathcal{U}$. See equation (\ref{input}). 
}

\begin{equation}
u = \{V, \gamma\}
\label{input}
\end{equation} where $V$ is the translational velocity and,  $\gamma$ is the steering angle of the front wheels. It is important to point out that a car-like vehicle has the non-holonomic constraint $\dot{y} \cos (\theta) - \dot{x} \sin (\theta) \equiv 0$, which limits its ability to move in arbitrary directions \cite{b3}, making difficult some tasks like parking manoeuvres.  

\begin{figure}
\includegraphics[width = \linewidth]{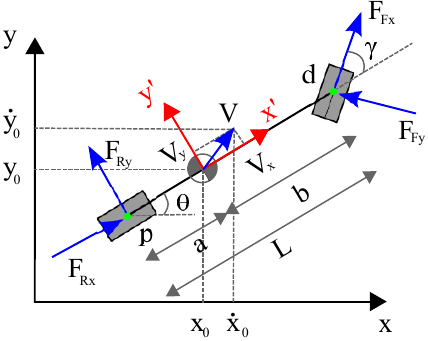}
\caption{\textcolor{black}{Forces diagram for the dynamic model of a car with rear and front traction. Where $F_{R_{x}}$, $F_{R_{y}}$ denote the force components for the rear tire with contact point in $p$. $F_{F_{x}}$, $F_{F_{y}}$ are the force components for the front tire with a contact point in $d$. $V$ is the velocity vector with origin in the center of mass and components in the body frame.}}
\label{figDynamicDiagram}
\end{figure}

\textcolor{black}{
\subsubsection{Dynamic model}
\label{subsec_dynamicModel}
For the task with a high-speed profile, we defined the STE according to equation (\ref{STEDynamicModel}). This model is obtained from the analysis of the diagram of forces shown in Figure \ref{figDynamicDiagram} where we presented a bicycle model of a car-like robot with traction in front an rear tires.}

\textcolor{black}{
\begin{equation}
f(x,u) = \begin{bmatrix}
\dot{\mathrm{x}} \\
\dot{\mathrm{y}} \\
\dot{\theta} \\
\dot{V}_{x} \\
\dot{V}_{y} \\
\dot{r}
\end{bmatrix} = \begin{bmatrix}
V_x\cos(\theta)-V_{y}\sin(\theta) \\
V_x\sin(\theta)+V_{y}\cos(\theta) \\
r \\ 
\frac{F_{Rx}-F_{Fy}\sin(\gamma)+F_{Fx}cos(\gamma)}{\mathrm{m}}+V_{y}\dot{\theta} \\
\frac{F_{Ry}+F_{Fy}\cos(\gamma)+F_{Fx}sin(\gamma)}{\mathrm{m}}-V_{x}\dot{\theta} \\
\frac{b(F_{Fy}cos(\gamma)+F_{Fx}sin(\gamma))-aF_{Ry}}{I}
\end{bmatrix}
\label{STEDynamicModel}
\end{equation}
}
\textcolor{black}{
Where $V_{x}$ and $V_{y}$ symbolizes the components for translation velocity and $r$ the rotational velocity about the z-axis. The physical parameters of the model are the mass ($\mathrm{m}$), the momentum of inertia about z-axis ($I$) and the distance to center of mass from the rear ($a$) and front ($b$) wheels. This values are taking from Table \ref{table_specificationRobot}.} The control input is defined according to \ref{inputsDynamicModel} 

\textcolor{black}{
\begin{equation}
    u = \{F_{Rx},F_{Fx}, \gamma, C_y \}
    \label{inputsDynamicModel}
\end{equation}
}

\textcolor{black}{
From the equation \ref{inputsDynamicModel} the forces $F_{R_{x}}$ and $F_{F_{x}}$ denotes the traction in the front and rear tires, $\gamma$ indicates the steering angle and $C_y$ the transverse stiffness coefficient that we estimate for the model car as $-1x10^{-3}$($\frac{N}{rad}$) with this coefficient 
we can calculated the lateral forces in the tires $F_{F_{y}}$ and $F_{R_{y}}$ agree with $F_{F_{y}}=C_{y}\alpha_{F}$ and $F_{F_{y}}=C_{y}\alpha_{R}$ where the slip angles $\alpha_{F}$ and $\alpha_{R}$ are approximated as $\alpha_{F}=\frac{V_{y}+rb}{V_{x}}-\gamma$ for the front tire and $\alpha_{R}=\frac{V_{y}-ar}{V_{x}}$ for the rear tire. 
}

\subsection{Rapidly-exploring random trees}

Sampling-based motion planning is a set of incremental sampling and searching algorithms. Their target is to avoid the explicit construction of the obstacle space, $\mathcal{X}_{obs}$, and to explore the state space, $\mathcal{X}_{space}$, with a sampling scheme \cite{b1}. The RRT is a special case of the family referred to as rapidly exploring Dense Trees (RDTs); these algorithms have a dense covering of the $\mathcal{X}_{space}$, and this feature allows them to be probabilistically complete. The goal is to build a topological graph or tree denoted by $\mathcal{G}(V,E)$ where each vertex is a state, and each edge $E$ is a path that connects two vertices inside the free space $\mathcal{X}_{free} =\mathcal{X}_{space}$ \textbackslash  $\mathcal{X}_{obs}$ \cite{b1}. Unlike PRM \cite{b1}, the RRT follows a single-query model, for each query defined by a initial $(x_{initial})$ and goal state $(x_{goal})$, the planner returns a tree and whether the query was successful, a branch of the tree is a path to the goal state. \textcolor{black}{In short, the RRT algorithm \cite{b1} follows the steps shown in Algorithm \ref{RRTmethod}}.

\begin{algorithm}[h]
\SetAlgoLined
\SetKwInOut{Input}{\textcolor{black}{Input}}\SetKwInOut{Output}{\textcolor{black}{Output}}\SetKwInOut{Parameters}{\textcolor{black}{Parameters}}

\Parameters{\textcolor{black}{PDF, STE, Metric function, Environment}}
\Input{\textcolor{black}{Query}}
\Output{\textcolor{black}{Path}}

\BlankLine
\textcolor{black}{Initialization}\\
\textcolor{black}{Vertex Selection Method (VSM)}\\
\textcolor{black}{Local Planing Method (LPM)}\\
\textcolor{black}{Graph update}\\ 
\textcolor{black}{Check for solution}\\
\textcolor{black}{Return to VSM}\\ 

\textbf{\textcolor{black}{Return}} \textcolor{black}{path}

\caption{\textcolor{black}{RRT methods}\label{RRTmethod}}
\end{algorithm}

The Vertex Selection Method (VSM) consist of generating a random vertex $x_{rand}\in \mathcal{X}_{space}$ between the proposed lower and upper state limits and finding the nearest existing vertex, according to a metric function $(\rho)$ then, if this vertex is an element of the free state space, $\mathcal{X}_{free}$, and it satisfies the conditions of Local Planning Method (LPM), this vertex is added to the tree as a new vertex, $x_{new} \in \mathcal{X}_{free}$. If the method reaches a solution it returns true and the algorithm finish, in other case, it returns to the VSM, this process is repeated for $k$ iterations.

\textcolor{black}{
\subsection{Metric functions}
To determine the similarity between two states, we propose the functions described in sub-section \ref{subsec_metricKinematic} for the kinematic model and a function for dynamic model in sub-section \ref{subsec_metricDynamic}. 
}
\textcolor{black}{
\subsubsection{Metric for kinematic model}
\label{subsec_metricKinematic}
For the kinematic model the similarity between two states $x_{1}=\{ \mathrm{x_1}, \mathrm{y_1}, \theta_{1} \}$ and $x_{2}=\{ \mathrm{x_2}, \mathrm{y_2}, \theta_{2} \}$ is a weighted and normalized sum defined by the equation (\ref{metrica1}).
}

\begin{equation}
\rho (x_1, x_2) = \omega_1 \cdot \rho_1 (x_1, x_2) + \omega_2 \cdot \rho_2 (x_1, x_2)
\label{metrica1}
\end{equation}

\noindent
where $\rho_1$ and $\rho_{2}$ determine the similarity between position and orientation, equations (\ref{rho1}) and (\ref{rho2}) respectively. \textcolor{black}{In equation (\ref{rho1}) the pairs ($x_{max},x_{min}$) and ($y_{max},y_{min}$) are the maximum and minimums values for the position states}. The variables $\omega_{1}$ and $\omega_{2}$ weigh the metrics for each of the proposed tasks and, they are statistically determined in every environment. In the cases presented with a kinematic model, task 1 and 4 $\omega_{1}=0.8$, $\omega_{2}=0.2$  and for task 2 and 3 $\omega_{1}=0.9$, $\omega_{2}=0.1$. \textcolor{black}{The weight's value change according to the relevance of being well-oriented or shortening distances to the objective for the current task.} 

\begin{equation}
\rho_1 = \left(\frac{(\mathrm{x_2}-\mathrm{x_1})^2+(\mathrm{y_2}-\mathrm{y_1})^2}{(x_{max}-x_{min})^2+(y_{max}-y_{min})^2}\right)^{\frac{1}{2}}
\label{rho1}
\end{equation}

\begin{equation}
\rho_2=\frac{\mathrm{min}(\mathrm{abs}(\theta_1-\theta_2), 2\pi-\mathrm{abs}(\theta_1-\theta_2))}{\pi}
\label{rho2}
\end{equation}

\textcolor{black}{
\subsubsection{Metric function for dynamic model}
\label{subsec_metricDynamic}
For the dynamic model the similarity between two states $$x_{1}=\{\mathrm{x}_{1},\mathrm{y}_{1},\theta_{1},V_{x_{1}},V_{y_{1}},r_{1}\}$$ and $$x_{2}=\{\mathrm{x}_{2},\mathrm{y}_{2},\theta_{2},V_{x_{2}},V_{y_{2}},r_{2}\}$$ is calculated in equation (\ref{eq_metricRoDRoK}) formed with the contribution of equation (\ref{eq_metricDynamic_k}) for the position and orientation states and equation (\ref{eq_metricKinematic_k}) for the dynamic variables.
}

\textcolor{black}{
\begin{equation}
\rho=\rho_{d}+\rho_{k}
\label{eq_metricRoDRoK}
\end{equation}
}

\textcolor{black}{
\begin{equation}
\begin{matrix}
\rho_k=k_{1}(\sqrt{(\mathrm{x}_{2}-\mathrm{x}_{1})^2})+ \\
k_{2}(\mathrm{min}(\mathrm{abs}(\theta_{1}-\theta_{2}),2\pi-\mathrm{abs}(\theta_{1}-\theta_{2})))
\end{matrix}    
\label{eq_metricDynamic_k}
\end{equation}
}

\textcolor{black}{
\begin{equation}
    \rho_d=k_{3}(\mathrm{abs}(V_{x_{1}}-V_{x_{2}})+\mathrm{abs}(V_{y_{1}}-V_{y_{2}}))+k_{4}(\mathrm{abs}(r_{1}-r_{2}))
    \label{eq_metricKinematic_k}
\end{equation}
Let that $k_{1,...,4}$ denoted a value to weighted the sum according to a specific characteristic of the problem.
}

\section{Custom sampling distribution}
\label{cpdfMethod}

In this section, we propose a methodology for constructing a custom probability density function (C-PDF) that will be used by the planner in order to improve its performance in similar tasks. The methodology consists of executing the RRT planner for an initial query that we denominated the construction query. If the construction query produces a successful path, then we collect specific data from the generated samples; otherwise, we discard the samples. We repeat this process until reaching $m$ samples to approximate a C-PDF. Finally, we use this function together with the rejection sampling method \cite{b2} as sampling source in a new set of queries improving the planner performance. In section \ref{collectingData}, we explain the details of this process taking the environment of task 1 as example and in section \ref{rejectionSampling}, we explain the details of the sampling method.      

\subsection{Collecting data}
\label{collectingData}

\begin{figure}[tb]
    \centering
    \includegraphics[width = 0.8\linewidth]{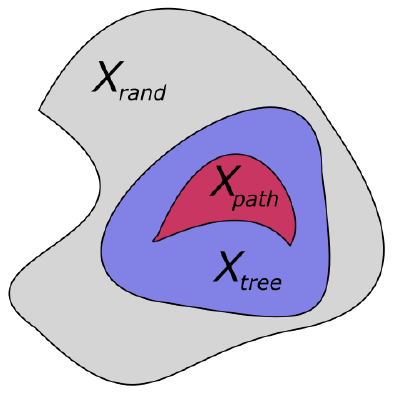}
    \caption{Venn diagram of sampling regions. In color gray the set of all random vertices $X_{rand}$, in color purple the sub-set of random vertices that contribute to generate the tree, $X_{tree}$ and in color red only the random vertices that contribute to a path that resolve the query, $X_{path}$.}
    \label{samplingRegionsl}
\end{figure}

\textcolor{black}{
Let us suppose that a uniform distribution, $\mathrm{U}$,} draws the samples during the tree construction. In ideal conditions, each random sample \textcolor{black}{($x_{rand}\in X_{rand}$)} should generate a new vertex ($x_{new}$) for the tree, $\mathcal{G}$. But for different reasons\textcolor{black}{, e.g., $x_{new}$ falls in the obstacle space or the new vertex is worst than the nearest vertex in terms of the metric function, only a reduced percentage of $x_{rand}$ samples produces a $x_{new}$ vertex. We call to the set of samples from $X_{rand}$ that generates a new vertex in the tree as $X_{tree}$.} \textcolor{black}{Note that $X_{rand}$ are random states and in the majority of the cases they do not correspond to the vertices in the tree; therefore, they do not contain the solution path. However, these random samples contain information for directing the tree growth.} So, for several queries, if one of them is successful then there is a branch in $\mathcal{G}$ that solves the query. This branch is the solution path and it is written as $\mathcal{P}$. $\mathcal{P}$ has $n$ elements and each one was drawn from a corresponding element in $X_{tree}$. 
So, we call to the sub-set of $X_{tree}$ that generated $\mathcal{P}$ as $X_{path}$. See Figure \ref{samplingRegionsl}. 

\begin{figure*}[tb]
\begin{subfigure}{}
\includegraphics[scale=0.20]{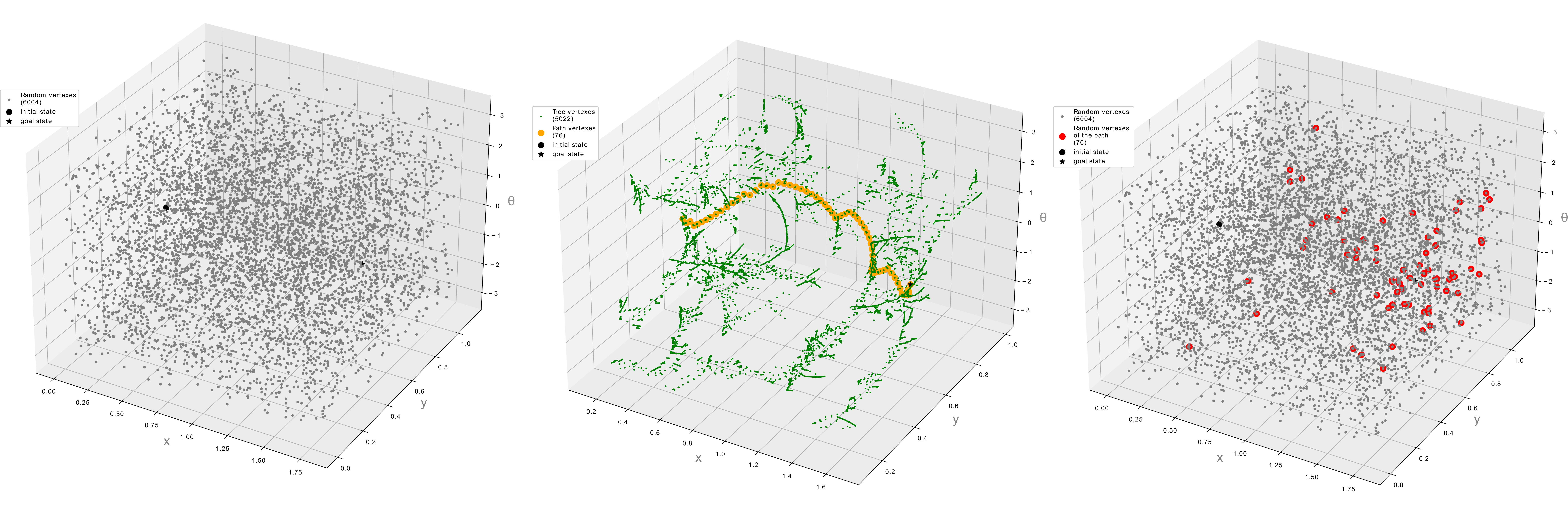}
\end{subfigure}
\vspace{0.01cm}
\caption{\textcolor{black}{The example shows the collecting process in the state space for a planning test in the environment for task 1. From left to right. All the random vertices in gray from $\mathrm{U}$ distribution. The tree of vertices is in color green, and the solution path is drawing with orange points. The random vertices are painted in gray and in red are the random vertices that generated the path, $X_{path}$. We can see how the samples  that direct the branching of the path follows a different distribution to U.}}
\label{samplingProcess}
\end{figure*}

By analyzing $X_{path}$, we can notice that these samples follow a different distribution compared with the original $\mathrm{U}$ and this trend continues for each new test for the same query. An example of this process can be seen in Figure \ref{samplingProcess}. \textcolor{black}{So, our hypothesis is that a solution path can be directly drawn from a distribution $\mathrm{C}$, which is different from $\mathrm{U}$.} Since $\mathrm{C}$ is unknown, our objective is inferring its shape by collecting a significant number of samples. To do this, we make several queries to the planner and we store the $X_{path}$ samples until $m$ elements are collected.

The collection process is summarized in Algorithm \ref{alg:collecting}. Where the output is the set $\mathcal{S}$ that denotes the set of $m$ samples that approximates $\mathrm{C}$. \textcolor{black}{Note that $\mathcal{S}$ is the set of $m$ samples, obtained from $X_{path}$, accumulated after repeating the construction query and, $X_{path}$ is the set of samples after a single query}. 
\textcolor{black}{Please note that the sets $\mathcal{S}$, $X_{rand}$, $X_{tree}$ and $X_{path}$ 
are obtained from a sampling process to guide the growth of the graph, and hence they might not be part of the graph, unlike the sets $\mathcal{P}$ and $X_{new}$ which belong to $\mathcal{G}$.} 
\par
Let $GENERATE\_RRT$ denote the RRT path planning described in Algorithm \ref{alg_rrt} and $RESOLVE\_QUERY$ the method for finding a path within $\mathcal{G}$. Please note that Algorithm 2 is proposed in \cite{b24}, we only add a simple modifications highlighted in gray to save and return the tree $\mathcal{G}$ together with the sets $\mathcal{X}_{rand}$ and $\mathcal{X}_{new}$. The line $5$ in $COLLECTING\_DATA$ uses the tree $\mathcal{G}$ along with the initial and goal state as inputs for the function $RESOLVE\_QUERY$ to find the path $\mathcal{P}$. \textcolor{black}{Finally in lines $7$ to $11$, if $\mathcal{P}$ is not an empty set, then we retrieve the samples $\mathcal{X}_{path}$ that build the path $\mathcal{P}$ in this case the function $GET\_PAIR$ receives the element $p_{i}$ together with the sets $\mathcal{X}_{rand}$ and $\mathcal{X}_{new}$ to seeks the correspondent sample, otherwise we do not add samples. The process continues until $m$ samples are reached.}\\

\begin{figure*}[tb]
\centering
\includegraphics[scale=0.29]{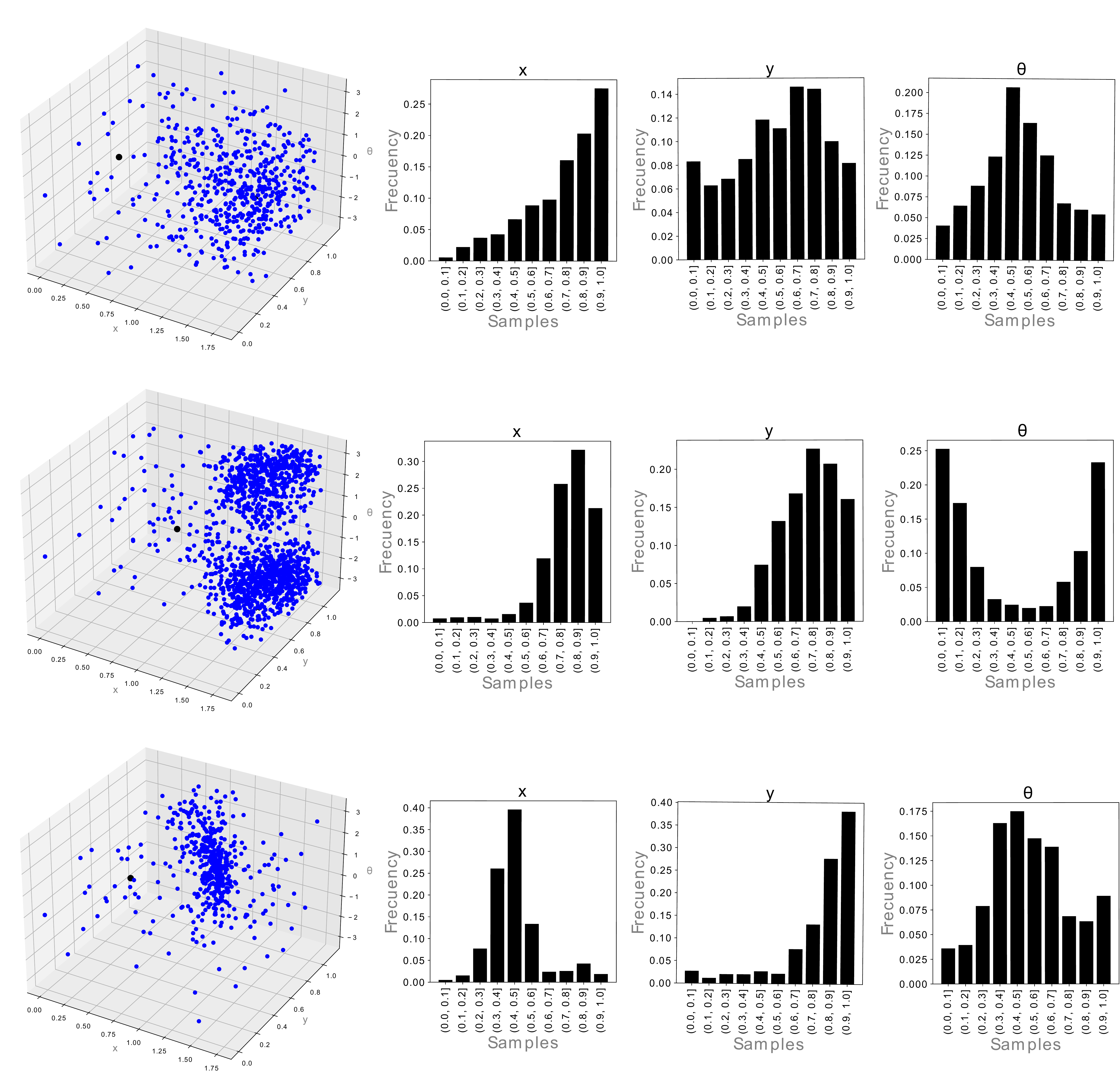}
\caption{\textcolor{black}{Set $\mathcal{S}$ and its respective histogram for the first three tasks. From left to right and top to bottom. For tasks 1 to 3 in color black, the set $\mathcal{S}$ in the state space and in black their histogram for independent variables $\{x,y,\theta\}$. We can see how the C distribution does not follow a U distribution because the tree does not need to explore the state space in its totality to reach a solution.}}
\label{allSamplesAndHisto}
\end{figure*}

\textcolor{black}{An open question is about how many samples, $m$, we need to approximate $\mathrm{C}$; for now $m$ is the number of collected samples after a fixed number of tests for a single o multiples construction queries. The complete information about the collecting process for each task is available in Table \ref{table_constructionQueryInformation}}. \textcolor{black}{The reader can see in Figure \ref{allSamplesAndHisto} the collecting process for one test in task 1}.

\begin{table}[]
\begin{tabular}{|c|c|c|c|c|}
\hline
\textbf{\textcolor{black}{Task}} & \textbf{\textcolor{black}{\begin{tabular}[c]{@{}c@{}}Type \\ of C. query\end{tabular}}} & \textbf{\textcolor{black}{\begin{tabular}[c]{@{}c@{}}Number \\ of tests\end{tabular}}} & \textbf{\textcolor{black}{\begin{tabular}[c]{@{}c@{}}Success \\ rate\end{tabular}}} & \textbf{\textcolor{black}{\begin{tabular}[c]{@{}c@{}}Samples\\ (m)\end{tabular}}} \\ \hline
\textcolor{black}{1}&\textcolor{black}{single(1)}&\textcolor{black}{20}&\textcolor{black}{0.4}&\textcolor{black}{541}\\ \hline
\textcolor{black}{2}&\textcolor{black}{single(1)}&\textcolor{black}{20}&\textcolor{black}{0.6}&\textcolor{black}{1464}\\ \hline
\textcolor{black}{3}&\textcolor{black}{single(1)}&\textcolor{black}{20}&\textcolor{black}{0.45}&\textcolor{black}{583}\\ \hline
\textcolor{black}{4}&\textcolor{black}{multiple(10)}&\textcolor{black}{10}&\textcolor{black}{0.18}&\textcolor{black}{2110}\\ \hline
\textcolor{black}{5}&\textcolor{black}{single(1)}&\textcolor{black}{10}&\textcolor{black}{0.7}&\textcolor{black}{134}\\ \hline
\end{tabular}
\caption{\textcolor{black}{Table with the information of the collecting process for each task. The column ``Type of C. query'' means the number of construction queries that we realized for the current task; the column ``Number of tests'' refer to how many times we repeat each query.} }
\label{table_constructionQueryInformation}
\end{table}

\begin{figure*}[h]
    \centering
    \includegraphics[scale=0.28]{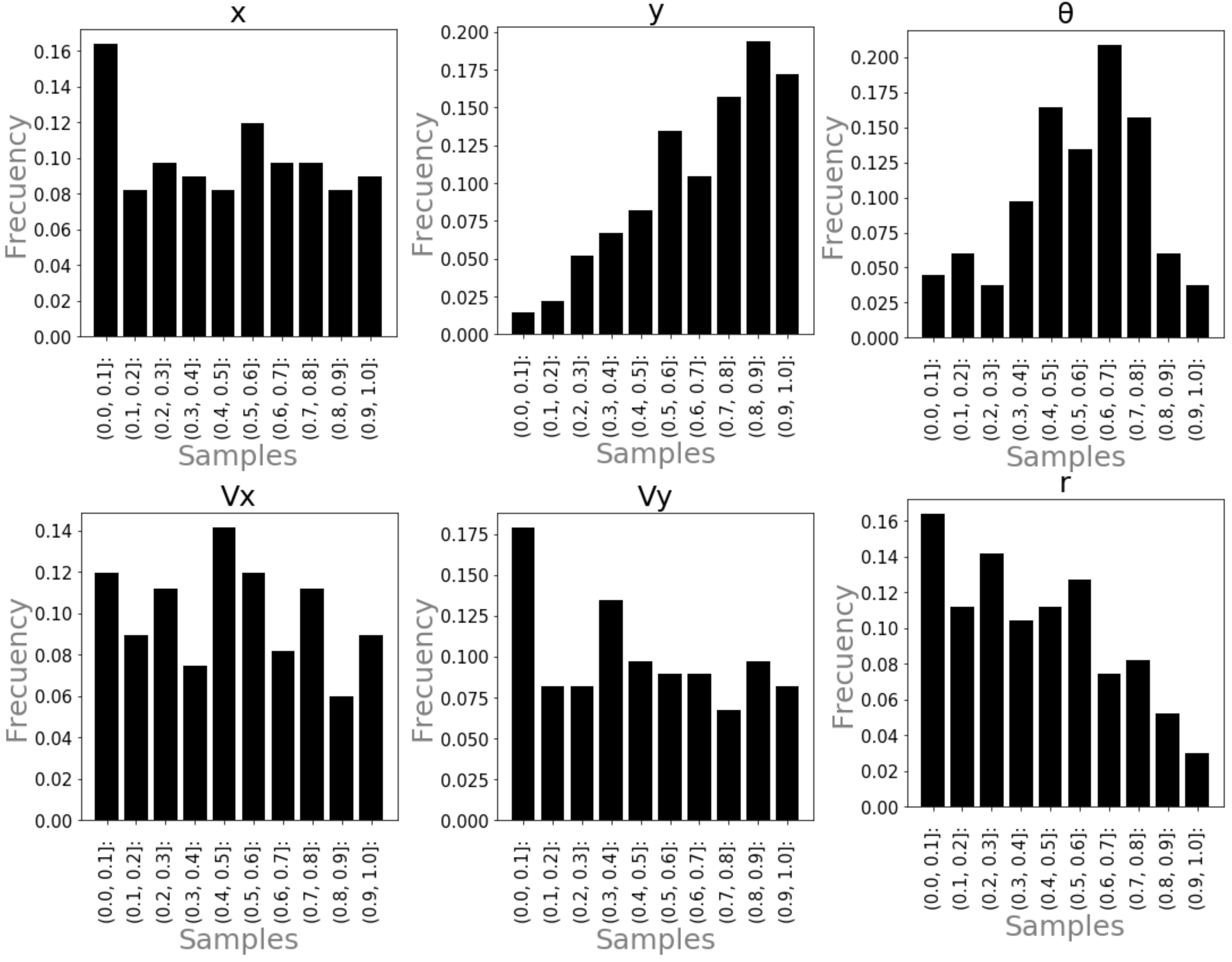}
    \caption{\textcolor{black}{Histogram for each state for task 5. We can notice how the random variables $x$, $V_{x}$ and $V_{y}$ do not show a clear difference regarding a uniform PDF while variables $y$, $\theta$ and $r$ follow a different distribution}}
    \label{samplesTaskDynamic}
\end{figure*}

\begin{algorithm}[tb]
\SetAlgoLined
\SetKwInOut{Input}{Input}\SetKwInOut{Output}{Output}\SetKwInOut{Parameters}{Parameters}
\Parameters{$k$, $\Delta_{t}$, $m$}
\Input{$x_{init}$, $x_{goal}$}
\Output{$\mathcal{S}$}
\BlankLine
\emph{$\mathcal{S}\leftarrow \emptyset$}\;
\While{$|\mathcal{S}| < m$}{
$\mathcal{G},\mathcal{X}_{rand},\mathcal{X}_{new}, X_{path} \leftarrow \emptyset$\;
$\mathcal{G}, \mathcal{X}_{rand}, \mathcal{X}_{new} \leftarrow$ \scalebox{0.85}{$GENERATE\_RRT(x_{init}, k, \Delta_{t})$}\;
$\mathcal{P}\leftarrow$ \scalebox{0.85}{$RESOLVE\_QUERY(x_{init}, x_{goal}, \mathcal{G})$}\;
	\If{$\mathcal{P} \neq \emptyset$}{
		\For{$i=0$ \KwTo $|\mathcal{P}|$}
		{
				$x_{rand}\leftarrow GET\_PAIR(\mathcal{X}_{rand},\mathcal{X}_{new},p_{i})$\;
				$X_{path}\leftarrow X_{path} \cup x_{rand}$\;
		}
	$\mathcal{S}\leftarrow \mathcal{S} \cup X_{path}$;
	}
}
\textbf{Return} $\mathcal{S}$
\caption{COLLECTING\_DATA()\label{IR}}
\label{alg:collecting}
\end{algorithm}

\begin{algorithm}[tb]
\SetAlgoLined
\BlankLine
$\mathcal{G}.init(x_{init})$\;
\For{$k=1$ \KwTo $K$}
{
	$x_{rand}\leftarrow RANDOM\_STATE()$\;
	$x_{near}\leftarrow NEAREST\_NEIGHBOR(x_{rand},\mathcal{G})$\;
	$u\leftarrow SELECT\_INPUT(x_{rand},x_{near})$\;
	$x_{new}\leftarrow NEW\_STATE(x_{near},u,\Delta_{t})$\;
	\colorbox{gray}{$\mathcal{X}_{rand} = \mathcal{X}_{rand} \cup x_{rand}$}\;
	\colorbox{gray}{$\mathcal{X}_{new} = \mathcal{X}_{new} \cup x_{new}$}\;
	$\mathcal{G}.add\_vertex(x_{new})$\;
	$\mathcal{G}.add\_edge(x_{near},x_{new},u)$\;
}
\colorbox{gray}{\textbf{Return} $\mathcal{G},\mathcal{X}_{rand},\mathcal{X}_{new}$}
\caption{GENERATE\_RRT($x_{init, K,\Delta_{t}}$)\label{IR}}
\label{alg_rrt}
\end{algorithm}

\subsection{Sampling data from the custom distribution}
\label{rejectionSampling}

Once we have several samples, then we use the rejection sampling method to generate new samples. Since $\mathrm{C}$ is unknown and it is only approximated, we call to the actual distribution C-PDF (custom PDF), where C-PDF is composed by the set $\mathcal{S}$. 
In order to find the C-PDF, we need express the states variables in a normalized 
sampling space. For this purpose, we use the equation (\ref{eq_sampleSpace}). 

\begin{equation}
\label{eq_sampleSpace}
r_{j,i}=\frac{\mathcal{S}_{j,i}-LS_{i}}{US_{i}-LS_{i}} 
\end{equation}

The sub-script $i$ iterates for each state variable, the sub-script $j$ iterates for each element of the set $\mathcal{S}$; $LS$ and $US$ are the lower and upper limits in the $X_{space}$. Applying the equation (\ref{eq_sampleSpace}) to each element, we obtain the sampling space for the state variables in each environment. 

\textcolor{black}{
There are two ways for drawing new samples: i) to consider the elements of the state as independent variables or ii) to include the correlation between them. In the experiment section, we test both cases. For the tasks with a low dimension state space, task 1, 2, 3, we express all the random variables as independent random variables. Where they are divided in ten bins to generate the histograms shown in Figure \ref{allSamplesAndHisto}.} 
\textcolor{black}{To improve the results in tasks with a more complex environment and with a larger search space. For the task 4 and 5, we consider the correlation between random variables}, see Figure \ref{histoT4} and \ref{samplesTaskDynamic}. In both cases, we use the rejection sampling method \cite{b2} to generate the C-PDF into the parameters of the planner. The idea of this method is sampling a target distribution $p(z)$ usually normalized from a more simple distribution $q(z)$ denominated proposal distribution. To apply the method we need to propose a scaling constant $c$ such that $p(z)\leq cq(z)$ i.e. the scaling proposal distribution covers the target distribution over the range of $z$. The next step is to generate a random number $z_{0}$ from the scaled proposal distribution and to propose a candidate sample $u_{0}$ in the interval $[0,cq(z_{0})]$ if the $u_{0}\leq p(z_{0})$ then the sample is accepted in other case is rejected. For a detailed explication the reader is refereed	to \cite{b2}. We implemented a discrete version of this method using a uniform distribution as proposal distribution and each one of the distributions formed by the variables $r_{i}$ as target distribution. \textcolor{black}{To differentiate the cases when we reconstruct the C-PDF with correlation we call it C*-PDF}  

\section{Experiments}
\label{evaluation}
\textcolor{black}{To evaluate the proposed methodology, we select environments that represent common tasks in the autonomous driving scene. First, we proposed three scenarios with equal dimensions but with different obstacle space distribution, tasks 1 to 3. In task 4, we proposed an environment with a larger search space and narrow passages. Finally, for task 5, we used a high-dimensional state space for a dynamic model. The proposed environments were designed for a test robot of scale 1:10 with the characteristics described in Table \ref{table_specificationRobot}. For all the experiments, the test queries are a set of random queries with the same $x_{goal}$ for which the C distribution was built, but a random $x_{init}.$} 

\begin{table}[tb]
\begin{center}
\begin{tabular}{|l|l|}
\hline
\multicolumn{2}{|c|}{Car-like robot} \\ \hline
       Height & $0.197$ \textsl{m}           \\ \hline
          Length & $0.39$ \textsl{m}           \\ \hline
          Width & $0.195$ \textsl{m}           \\ \hline
          Distance between axis($L$) & $0.255$ \textsl{m}           \\ \hline
          Rear distance to center of gravity & $0.14$ \textsl{m}           \\ \hline
          Front distance to center of gravity & $0.115$ \textsl{m}          \\ \hline
          Mass($\mathrm{m}$) & $0.2891$\textsl{kg}          \\ \hline
          Moment of inertia ($I_{z}$) & $4.7x10^{-3}\textsl{kg}\textsl{m}^2$          \\ \hline
\end{tabular}
\end{center}
\caption{Test robot specifications}
\label{table_specificationRobot}
\end{table}

We describe in sub-section \ref{proposedtasks} the generality of proposed environments. In sub-section \ref{numericalResults}, we describe the performance metrics. In the section \ref{experimentsAnalysis}, we expose the comparative results for tasks 1 to 3 between the U-PDF and the C-PDF and the RRT goal bias version (GB-PDF) for a group of ten random queries. \textcolor{black}{For task 4 and task 5, we expose the comparative for U-PDF, GB-PDF, C-PDF and C*-PDF for a group of three and five random queries respectively.}

\begin{figure}[tb]
\centering
\includegraphics[scale=0.3]{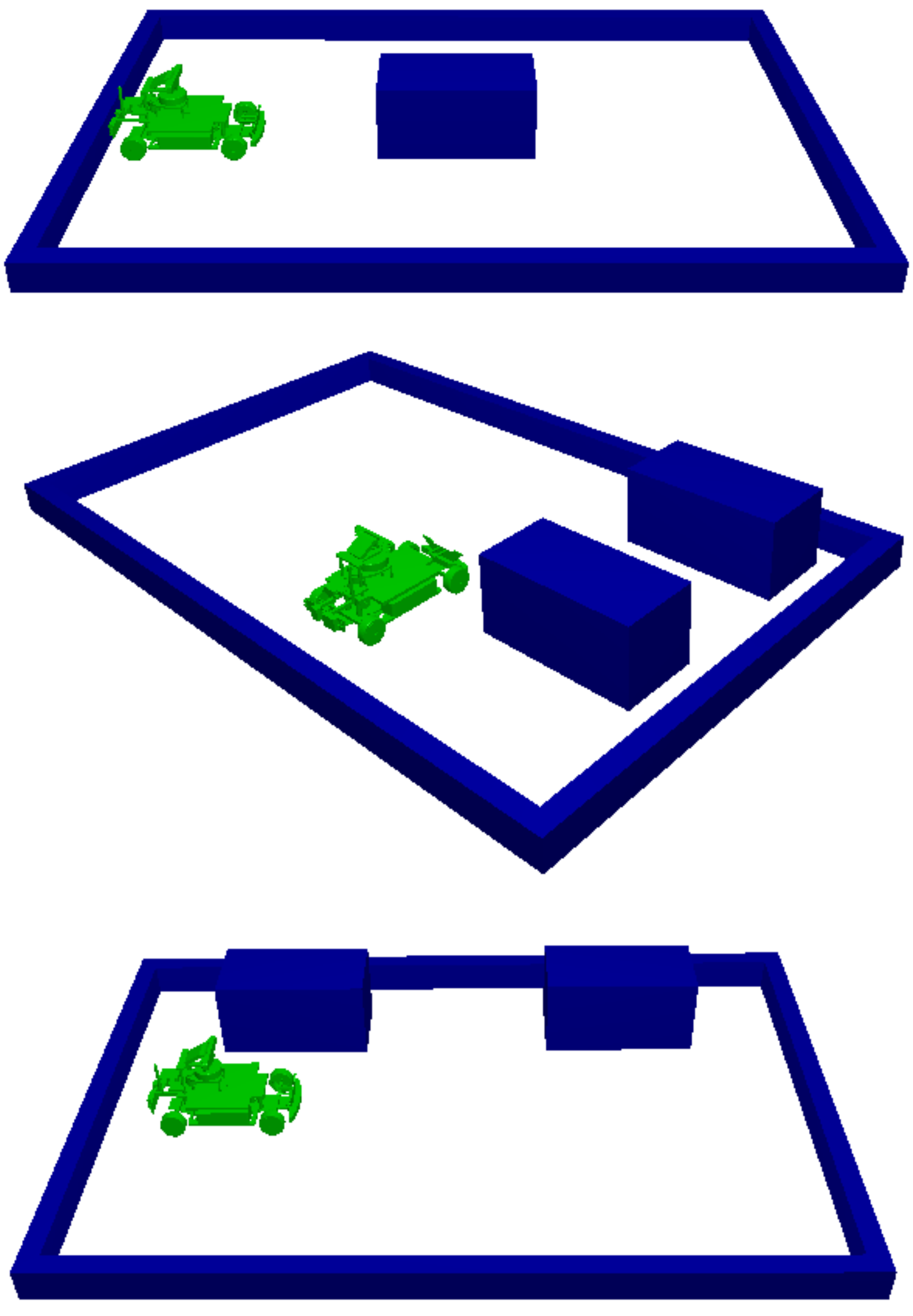}
\caption{Representation of the proposed tasks using Motion Planning Library \cite{b11}. From top to bottom; avoiding a static obstacle, parallel parking and line parking, respectively. In color black the obstacles in color white the free workspace and color green the robot.}
\label{proposedTasks}
\end{figure}

\newpage
\subsection{Proposed tasks}
\label{proposedtasks}

\begin{figure}[tb]
\centering
\includegraphics[scale=0.5]{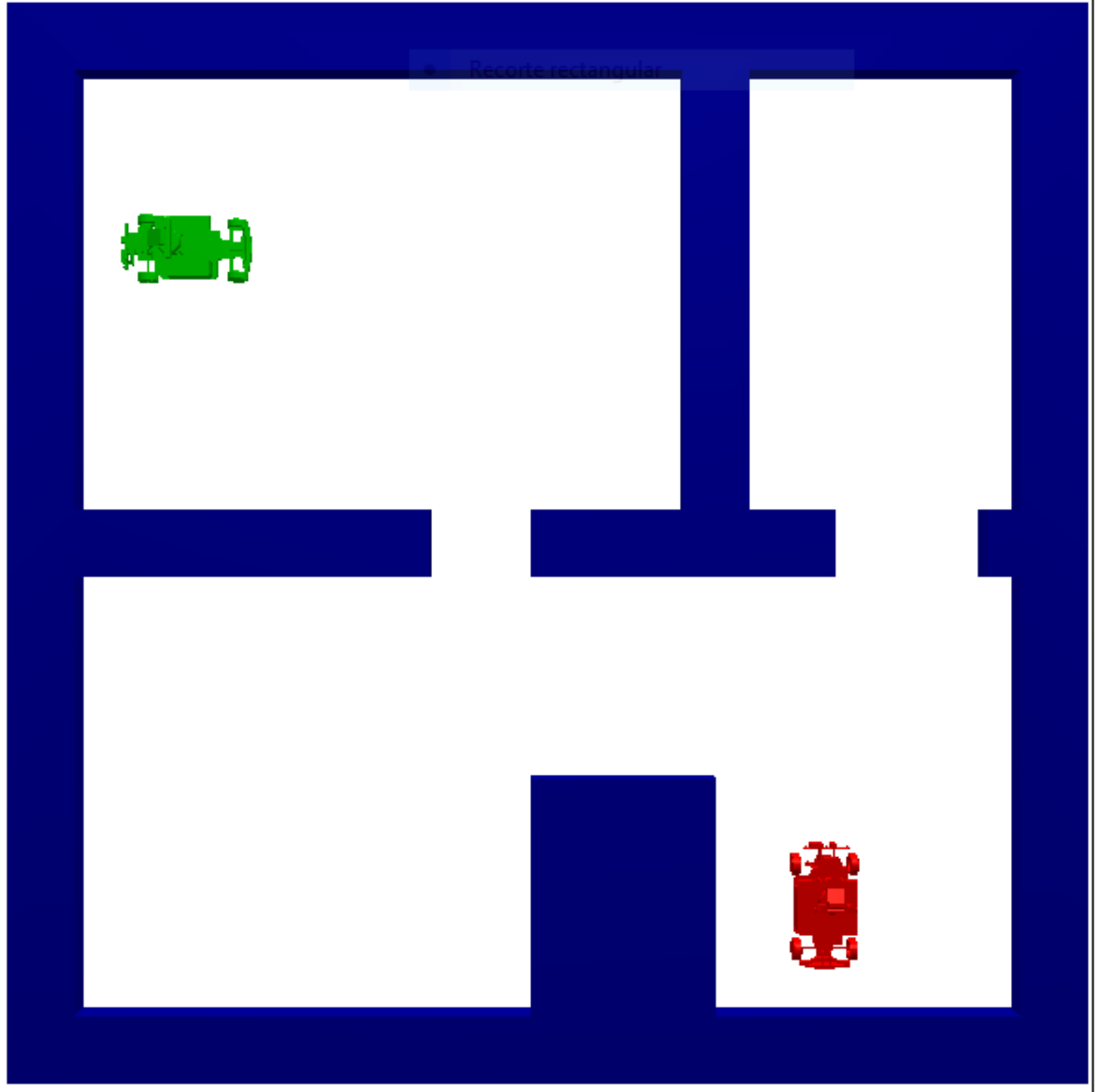}
\caption{\textcolor{black}{The task with narrow passages. The robot in color green represents the initial state and the red color final state. This task represents a larger search space where the robot should cross a narrow passage to solve the query. A narrow passage is a classic problem for SBMP algorithms.}}
\label{complexTask}
\end{figure}

\begin{figure}
    \centering
    \includegraphics[scale=0.7]{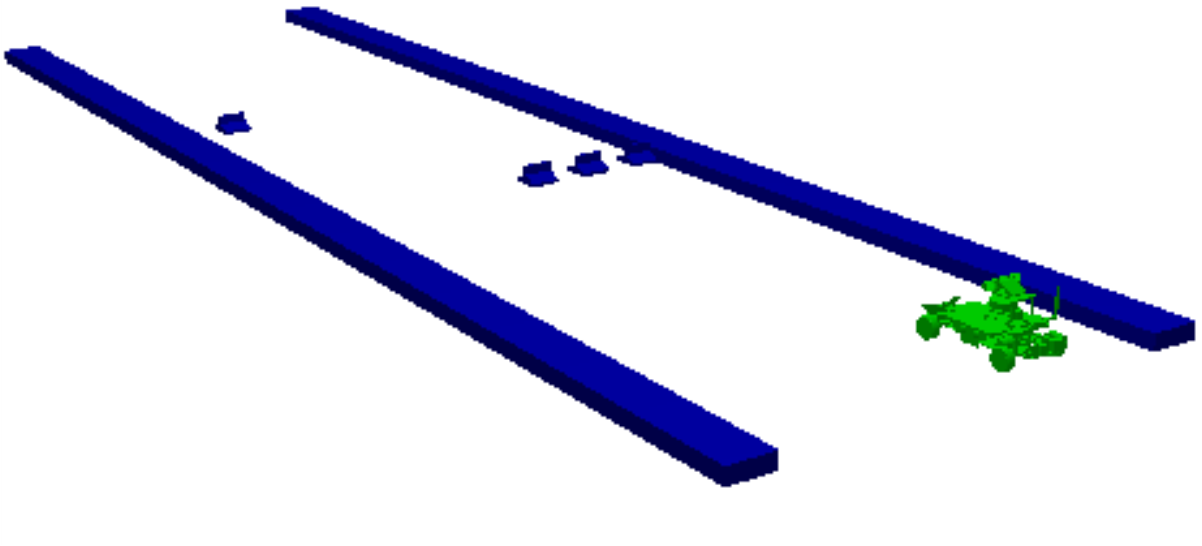}
    \caption{Task 5. In this task the car-like robot should start and finish with a high speed about $0.5m/s$ while it must avoid static obstacles in the road.}
    \label{fig_task5}
\end{figure}

\begin{figure}
\centering
\includegraphics[width = \linewidth]{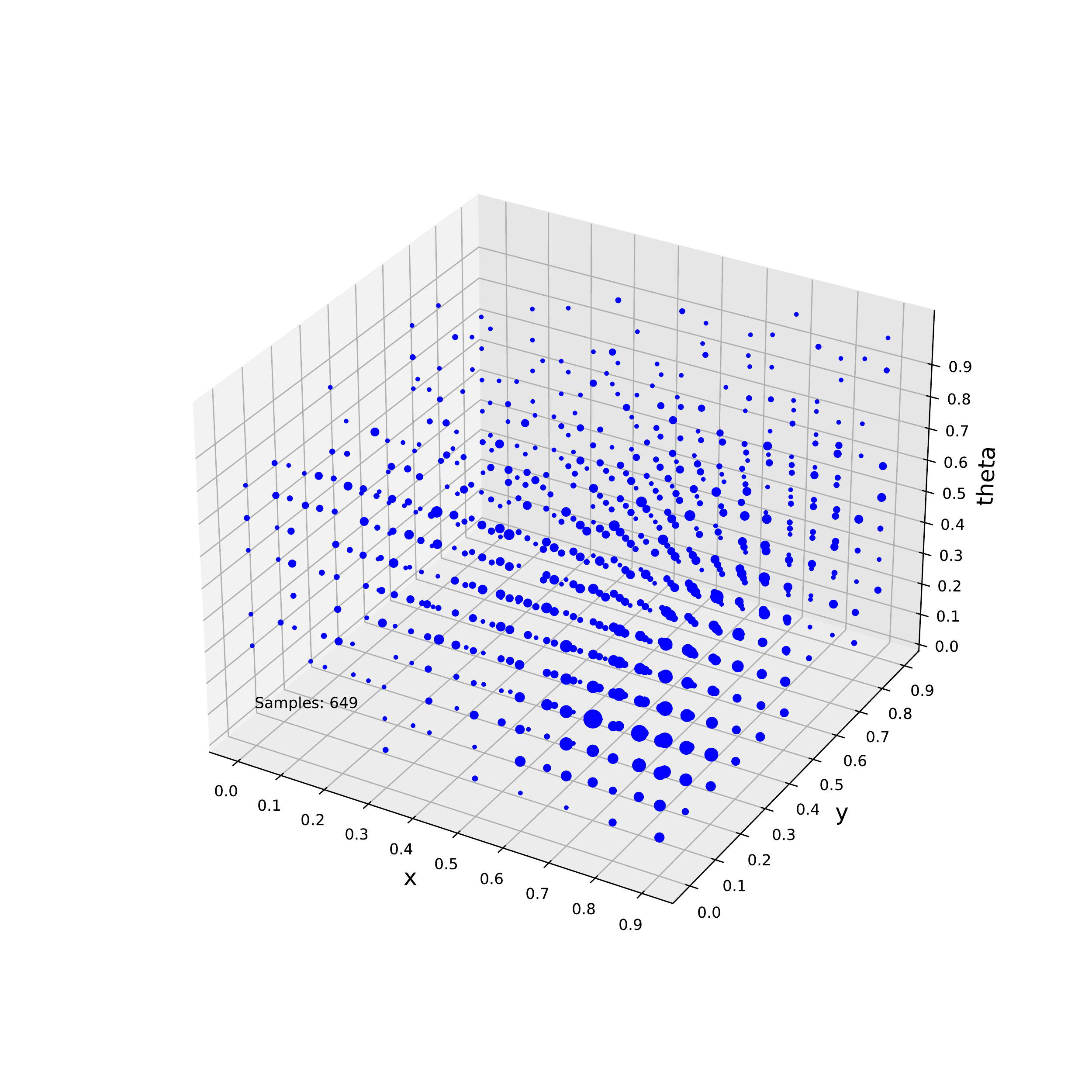}
\caption{ \textcolor{black}{For task 4, the histogram for the custom probability density function. In color black, the spheres which radius represent the number of samples for an interval. This is a better histogram representation for the C distribution because the figure considers the correlation between random variables.}}
\label{histoT4}
\end{figure}

All the tasks are defined in a $\mathcal{W}$, such that it contains an obstacle region described as $\mathcal{O}\subset \mathcal{W}$. For the first three tasks, see Figure \ref{proposedTasks}, the lower and upper world coordinates for the $\mathcal{X}_{space}$ are defined as $X_{LS}=\{ 0.0, 0.0,-\pi \}$ for Lower State ($LS$) and $X_{US}=\{ 1.8, 1.125, \pi \}$ for Upper State ($US$). For the fourth task, see Figure \ref{complexTask}, $X_{LS}=\{ 0.0, 0.0,-\pi \}$ for $LS$ and $X_{US}=\{ 3.1, 3.1, \pi \}$ for $US$. \textcolor{black}{For the fifth task, see Figure \ref{fig_task5},  with a dynamic model $X_{LS}=\{ 0.0, 0.0,-\pi,-0.1,-0.1,-0.5 \}$ for $LS$ and $X_{US}=$ $\{ 1.2$, $5.0, \pi,0.5,0.5,0.5 \}$ for $US$.} 

Based on the previous scenarios the proposed tasks are:    

\begin{itemize}
\item Task 1: Avoiding a static obstacle. The goal in this task is to avoid a static obstacle that obstructs the route of the car.
\item Task 2: Parallel parking: Parking in a lot with dimensions $[0.40m, 0.30m]$ and between two rectangles. 
\item Task 3: Line parking. Parking in a lot with dimensions $[0.50m, 0.20m]$ and between two rectangular obstacles placed longitudinally.
\item Task 4: Narrow passages. Find a route in map with narrow passages.
\textcolor{black}{
\item Task 5: High speed obstacle avoidance. To avoid static obstacles while the robot advances with a high-speed profile. In this task, the dynamic model is applied.}  
\end{itemize}

\noindent

\noindent
Each experiment with the kinematic model used a set of discrete control inputs formed by three sub-set inputs denoted as low-speed, high-speed and reverse-speed formed a total of $58$ discrete inputs $19$ by each sub-set, see table \ref{tablaEntradas}. For example, the input number 2 is the pair $u_{2}=\{0.05,-45\}$, the input $u_{25}=\{0.01,-25\}$ and in the same way by the consecutive inputs.

\begin{table}[tb]
\begin{center}
\begin{tabular}{|l|l|l|}
\hline
\textbf{$u_{1,2,.. ,58}$}&\textbf{$V(\textsl{m}/\textsl{s})$}  &\textbf{$\gamma(\textsl{deg})$} \\ \hline
Stop & $0$ & $0$ \\ \hline
Hight velocity & $0.05$ & $\{-45, -40, -35,...,40,45\}$ \\ \hline
Low velocity & $0.01$ & $\{-45, -40, -35,...,40,45\}$ \\ \hline
Reverse & $-0.01$ & $\{-45, -40, -35,...,40,45\}$ \\ \hline
\end{tabular}
\caption{Set of 56 discrete control inputs for each proposed task. Divided into four sub-categories of velocity.}
\label{tablaEntradas}
\end{center}
\end{table}

\textcolor{black}{
In task 5, we used in addition to the stop input three input sets conformed by 19 inputs each one; low traction forces to move forward and backward and high traction forces to move only forward as shown in Table \ref{table_inputsDynamics}.    
}

\begin{table*}[]
\begin{center}
\scalebox{1.0}{
\begin{tabular}{|l|l|l|l|l|}
\hline
\textcolor{black}{\textbf{$u_{1,2,.. ,58}$}} & \textcolor{black}{\textbf{$F_{F_{x}}(\textsl{N})$}} & \textcolor{black}{\textbf{$F_{R_{x}}(\textsl{N})$}} & \textcolor{black}{\textbf{$\gamma (\textsl{deg})$}} & \textcolor{black}{\textbf{$C (\textsl{N/rad)}$}} \\ \hline
\textcolor{black}{Stop} & \textcolor{black}{$0.00$} & \textcolor{black}{$0.00$} & \textcolor{black}{$0$} & \textcolor{black}{$0.001$} \\ \hline
\textcolor{black}{High velocity} & \textcolor{black}{$0.005$} & \textcolor{black}{$0.005$} & \textcolor{black}{$\{-45, -40, -35,...,40,45\}$} & \textcolor{black}{$0.001$} \\ \hline
\textcolor{black}{Low velocity} & \textcolor{black}{$0.001$} & \textcolor{black}{$0.001$} & \textcolor{black}{$\{-45, -40, -35,...,40,45\}$} & \textcolor{black}{$0.001$} \\ \hline
\textcolor{black}{Reverse} & \textcolor{black}{$-0.001$} & \textcolor{black}{$-0.001$} & \textcolor{black}{$\{-45, -40, -35,...,40,45\}$} & \textcolor{black}{$0.001$} \\ \hline
\end{tabular}
}
\caption{\textcolor{black}{Set of 56 discrete control inputs for the dynamic model of a car-like robot}}
\label{table_inputsDynamics}
\end{center}
\end{table*}

\subsection{Performance metrics}
\label{numericalResults}
To evaluate the performance of our method we utilized the next four metrics:

\begin{itemize}
\item[1.] Tree density. The number of vertices of the tree. 
\item[2.] Connectivity percentage. The percentage of iterations that produces and connects a new vertex. 
\item[3.] Vertices in the path. The amount of vertices in the solution path.
\item[4.] Success rate. The percentage of queries that find a solution within a limit of $10, 000$ iterations.
\end{itemize}  

\begin{figure*}[tb]
\centering
\includegraphics[scale=0.35]{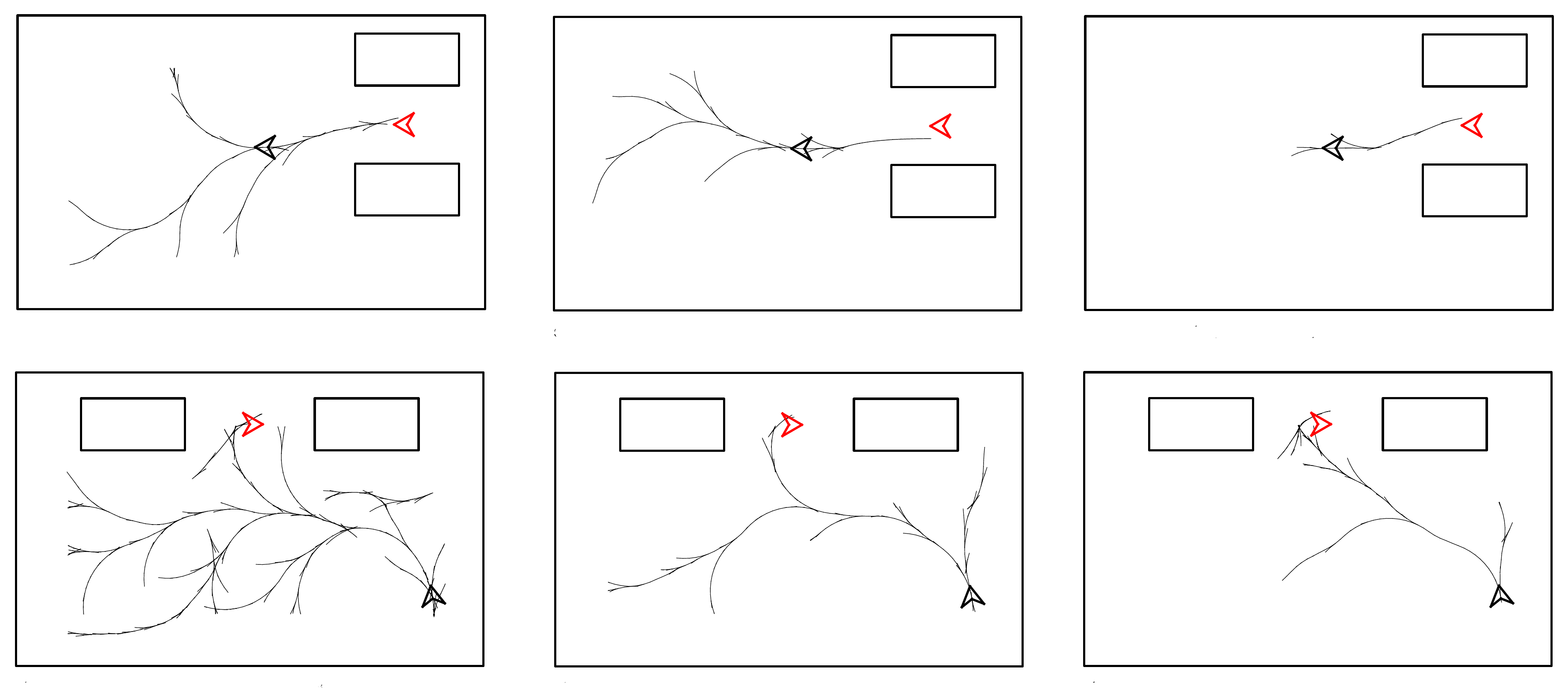}
\caption{\textcolor{black}{Tree results for tasks 2 and 3. From left to right the figure shows the resulting trees for a U-PDF, GB-PDF, and  C-PDF.  All the figures show a tree that includes a branch to resolve the query, but the tree for C-PDF resolves the query with a less dense tree.}}
\label{examplesTrees}
\end{figure*}

\begin{figure*}[]
\centering
\includegraphics[scale=0.45]{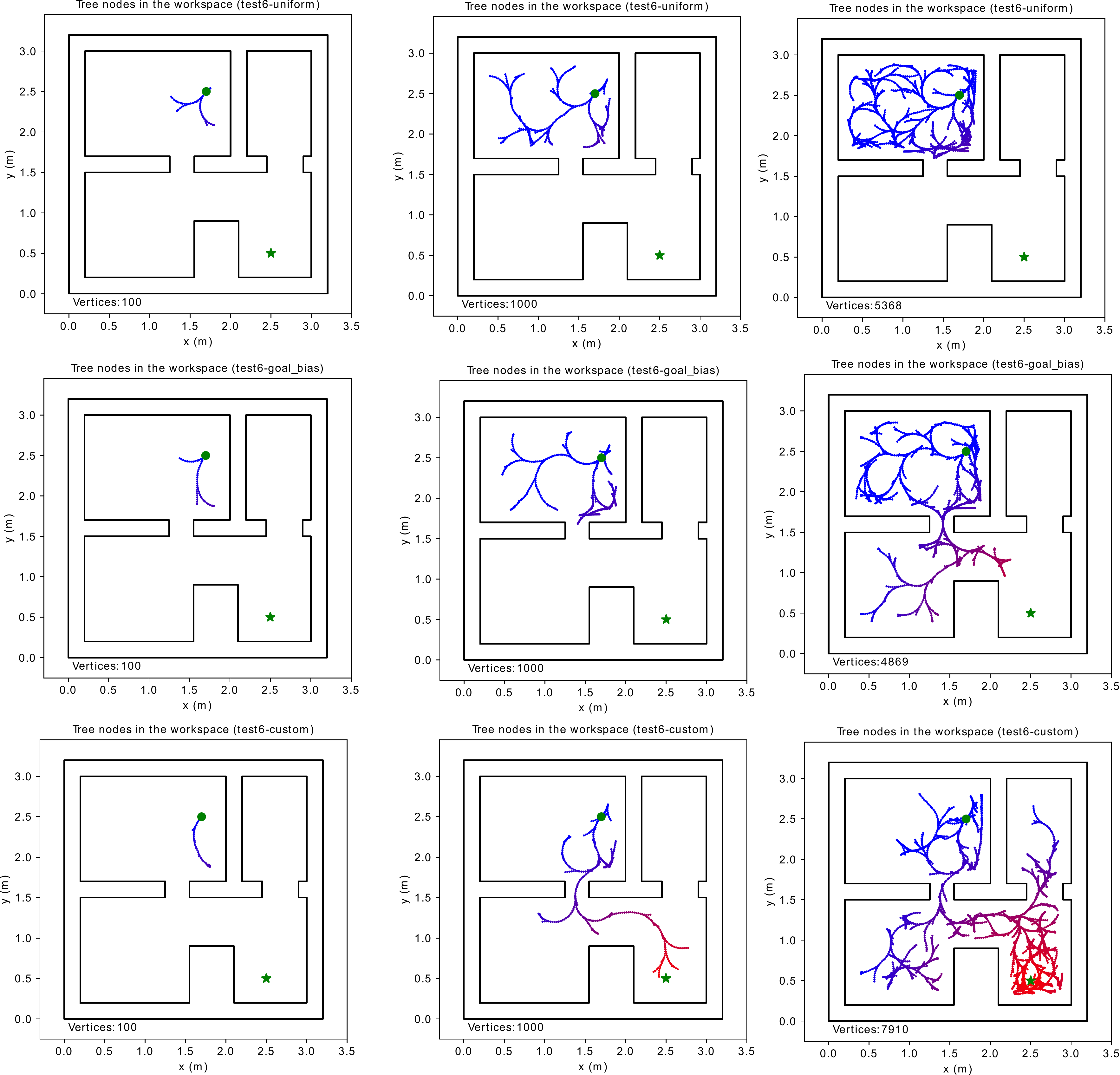}
\caption{Tree growth for the environment with narrow passages. From top to bottom, task 4 with a U-PDF, GB-PDF and C*-PDF. From left to right graph  for $100$, $1 000$ and all the vertices for each case with $10000$ iterations. In black are the vertices near to the initial state and in red color vertices close to the goal state. Figure best seen in color.}
\label{fig11}
\end{figure*}

\subsection{Experiments analysis}
\label{experimentsAnalysis}
In Table \ref{tableTaskRQ}, we present the results for tasks 1 to 3 using a U-PDF, GB-PDF, C-PDF and C*-PDF. \textcolor{black}{In Table \ref{tableTaskDynamic}, we present the results for task 5 for using U-PDF, GB-PDF, C-PDF and C*-PDF.}

\begin{table*}[]
\centering
\begin{tabular}{|c|l|l|l|l|l|}
\hline
\multicolumn{6}{|c|}{\textbf{Random query for task 1 to 3}}\\ \hline
\textbf{\begin{tabular}[c]{@{}c@{}}Sampling\\ function\end{tabular}}    & \textbf{\begin{tabular}[c]{@{}c@{}}Tree\\ density\end{tabular}} & \textbf{\begin{tabular}[c]{@{}c@{}}Connectivity\\ percentage\end{tabular}} & \textbf{\begin{tabular}[c]{@{}c@{}}Vertices in \\ path\end{tabular}} & \textbf{\begin{tabular}[c]{@{}c@{}}Path in \\ meters\end{tabular}} & \textbf{\begin{tabular}[c]{@{}c@{}}Success\\ rate\end{tabular}} \\ \hline
\multicolumn{6}{|c|}{\textbf{Task 1}}\\ \hline
\textbf{U-PDF}&                                                     $7041.60$&\textbf{0.83}&\textbf{73.60}&$1.02$&$0.26$\\ \hline
\textbf{GB-PDF}&                                                    \textbf{3718.40}&$0.59$&$85.35$&$1.06$&$0.53$\\ \hline
\textbf{C-PDF}&                                                     $4152.63$&$0.79$&$78.62$&\textbf{0.92}&\textbf{0.60}\\ \hline
\multicolumn{6}{|c|}{\textbf{Task 2}}\\ \hline
\textbf{U-PDF}&                                                     $4638.13$&\textbf{0.65}&\textbf{164.04}&$1.43$&$0.53$\\ \hline
\textbf{GB-PDF}
&$2543.70$&$0.47$&$165.50$&\textbf{1.15}&$0.66$\\ \hline
\textbf{C-PDF}
&\textbf{570.26}&$0.38$&$216.45$&$1.51$&\textbf{0.96}\\ \hline
\multicolumn{6}{|c|}{\textbf{Task 3}}\\ \hline
\textbf{U-PDF}&
$4665.66$&\textbf{0.65}&\textbf{114.43}&\textbf{1.33}&$0.50$\\ \hline
\textbf{GB-PDF}&
$2493.30$&$0.49$&$144.74$&$1.43$&$0.66$\\ \hline
\textbf{C-PDF}&                                                     \textbf{1224.16}&$0.48$&$189.45$&$1.61$&\textbf{0.96}\\ \hline
\end{tabular}
\caption{The table shows the comparative results between U-PDF, GB-PDF and C-PDF for a group of ten random queries for task 1 to 3. For each task the best results are highlighted.}
\label{tableTaskRQ}
\end{table*}
\begin{table*}[]
\centering
\scalebox{1.0}{
\begin{tabular}{|c|l|l|l|l|l|}
\hline
\multicolumn{6}{|c|}{\textbf{Random query for task 4}}\\ \hline
\textbf{\begin{tabular}[c]{@{}c@{}}Sampling\\ function\end{tabular}}    & \textbf{\begin{tabular}[c]{@{}c@{}}Tree\\ density\end{tabular}} & \textbf{\begin{tabular}[c]{@{}c@{}}Connectivity\\ percentage\end{tabular}} & \textbf{\begin{tabular}[c]{@{}c@{}}Vertices in \\ path\end{tabular}} & \textbf{\begin{tabular}[c]{@{}c@{}}Path in \\ meters\end{tabular}} & \textbf{\begin{tabular}[c]{@{}c@{}}Success\\ rate\end{tabular}} \\ \hline
\multicolumn{6}{|c|}{\textbf{Task 4}}                                  \\ \hline
\textbf{U-PDF}&                                                     $6265.13$&$0.70$&\textbf{110.00}&\textbf{2.65}&$0.20$\\ \hline
\textbf{GB-PDF}&                                                    \textbf{4064.86}&$0.50$&$202.20$&$3.29$&$0.36$\\ \hline
\textbf{\textcolor{black}{C-PDF}}&
\textcolor{black}{$5953.66$}&\textbf{\textcolor{black}{0.73}}&\textcolor{black}{$159.37$}&\textcolor{black}{$2.66$}&\textcolor{black}{0.26}\\ \hline
\textbf{C* -PDF}&
$5030.46$&0.72&$151.05$&$2.80$&\textbf{0.46}\\ \hline
\end{tabular}
}
\caption{The table shows the comparative results between U-PDF, GB-PDF and C-PDF for a group of ten random queries for task 4. For each task the best results are highlighted.}
\label{tableTaskRQ4}
\end{table*}

\begin{table*}[]
\centering
\scalebox{1.0}{
\begin{tabular}{|c|l|l|l|l|l|}
\hline
\multicolumn{6}{|c|}{\textbf{Random query for task 5}}\\ \hline
\textbf{\begin{tabular}[c]{@{}c@{}}Sampling\\ function\end{tabular}}    & \textbf{\begin{tabular}[c]{@{}c@{}}Tree\\ density\end{tabular}} & \textbf{\begin{tabular}[c]{@{}c@{}}Connectivity\\ percentage\end{tabular}} & \textbf{\begin{tabular}[c]{@{}c@{}}Vertices in \\ path\end{tabular}} & \textbf{\begin{tabular}[c]{@{}c@{}}Path in \\ meters\end{tabular}} & \textbf{\begin{tabular}[c]{@{}c@{}}Success\\ rate\end{tabular}} \\ \hline
\multicolumn{6}{|c|}{\textcolor{black}{\textbf{Task 5}}}                                  \\ \hline
\textcolor{black}{\textbf{U-PDF}}&                                                     \textcolor{black}{359.46}& \textcolor{black}{7.87}&\textcolor{black}{21.95}&\textcolor{black}{$21.95$}&\textcolor{black}{0.46}\\ \hline
\textcolor{black}{\textbf{GB-PDF}}&                                                    \textcolor{black}{310.59}&\textcolor{black}{6.45}&\textcolor{black}{\textbf{21.66}}&\textcolor{black}{$21.66$}&\textcolor{black}{0.33}\\ \hline
\textbf{\textcolor{black}{C-PDF}}&
\textcolor{black}{315.13}&\textcolor{black}{\textbf{14.88}}&\textcolor{black}{22.23}&\textcolor{black}{$22.23$}&\textcolor{black}{0.60}\\ \hline
\textcolor{black}{\textbf{C* -PDF}}&
\textcolor{black}{\textbf{201.86}}&\textcolor{black}{13.52}&\textcolor{black}{22.70}&\textcolor{black}{$22.70$}&\textcolor{black}{\textbf{0.66}}\\ \hline
\end{tabular}
}
\caption{The table shows the average results for U-PDF, GB-PDF, C-PDF and C*-PDF for a group of five random queries obtained from a gaussian distribution with $\mu$ around the initial state of the construction query.}
\label{tableTaskDynamic}
\end{table*}


\subsubsection{Random query results for task 1 to 3}

We generated a set of random $x_{init}$, while the $x_{goal}$ is fixed. We show the comparative results for the random queries in each scenario using the U-PDF, GB-PDF and C-PDF in the Table \ref{tableTaskRQ}. We summarized the result below:    

\begin{itemize}
\item The C-PDF increases the success rate by +$0.95$\% with respect to U-PDF and +$37.70$\% with respect to GB-PDF.    
\item The density of the tree decreases by $-63.61$ \% with respect to U-PDF and -$32.07$\% with respect to the goal bias version.  
\end{itemize}

This results imply less dense tree, therefore, less computational time when the nearest node is computed. We also have a considerable increment in the probability of success when making a query. This advantages are important because a robot in real operation need to plan in a short time. On the other hand, we have some negative consequence when using the C-PDF:

\begin{itemize}
\item Connectivity is reduced by -$22.53$\% with respect to U-PDF but it is increased +$7.84$\% with respect to GB-PDF. 
\item The number of the vertex in the path is increased by +$37.62$\% with respect to U-PDF and +$22.47$\% with respect to goal bias RRT version. 
\end{itemize}

The decrease in the percentage of connectivity means that more iterations do not add nodes to the tree. Although the decrease is not significant, more precise modelling of the distribution could reduce this adverse effect. On the other hand, the increase in the number of vertices in the path is the result of having an increase in branches near the goal state, which implies that its impact on the distance in meters does not have a significant increase. Examples of the tree for each distribution can be seen in Figure \ref{examplesTrees}.

\subsubsection{Random query results for task 4}

\noindent
For the narrow passage task (task 4), we include two versions of the custom PDF; the version without correlation, C-PDF, and the version with correlation, C*-PDF. The reader can see the complete results in Table \ref{tableTaskRQ4}. Table \ref{tableTaskRQ4} shows that the bests results are obtained when we use the C*-PDF because it is a better approximation for the custom function C. The loss of connectivity and the increase of the number of vertices in the path are natural consequences of having a skewed distribution due to the loss of variety in the generation of random vertices. However, for the connectivity is important to do not have an excessive loss in order to reduce the computing time. The results show that our method has better connectivity than the goal bias version, although it is worst than the U-PDF but this increase is not significant since we have sparse trees. For the path measurement, we only consider, by obvious reason the successful queries. In Table \ref{tableTaskRQ}, we present the path in terms of the number of vertices and path distance in meters. In general fewer vertices imply fewer control actions, therefore, energy savings. Our C*-PDF presents a considerable increment with respect to a U-PDF (+$37$\%) in terms of vertices but a low increment in terms of meters (+$6$\%) which means that the founding path has redundant actions, we consider that an optimization process can reduce the number of vertices.  

\noindent
In Figure \ref{fig11}, we can see the gradual tree growth. Please observe, that when we use a C*-PDF, we can reach the goal region in fewer vertices. We can see how 1000 vertices are enough to have a branch in the goal region. With the uniform and the GB-PDF, the tree has no branches outside the first quadrant of the room. Moreover, for U-PDF, after 10,000 iterations, any vertex could get out of the first quadrant. For the GB-PDF, after 10,000 iterations, the tree has branches in the first and third quadrant, but only a few branches begin to approach to the goal region. While with the C*-PDF, we have a branch in every quadrant and an important vertices density in the goal region, therefore a greater probability of finding a solution. In other words, the C*-PDF explodes the workspace more efficiently than the uniform and goal bias versions. That is an important characteristic in any search method.

\subsubsection{Random query results for task 5}
In Table \ref{tableTaskDynamic}, we show the results for the dynamic tests. We can observe how the trend to find a path solution in a limited number of iterations increases when we use a C-PDF, about +0.14 concerning a U-PDF and +0.27 concerning a GB-PDF. This increment is more noticeable for C*-PDF. For this task, the tree density remains without major changes for U-PDF, GB-PDF and C-PDF but with a notable reduction for C*-PDF. We can explain this issue due to the path that resolves the query needs to explore the entire workspace, but the C*-PDF reduces the search space. Another outstanding fact is the significant increment in the connectivity percentage for the C-PDF and C*-PDF, about double with respect to the other two distributions. We believe that a custom PDF improves the connectivity in space states, where the metric is difficult to be obtained, due to the lower percentage of connectivity. For this experiment, we can say that a custom PDF provides feasible directions of growth. In the same way that tasks 1 to 3, we can observe that a C-PDF does not create paths that improve or impair the solution in terms of the number of vertices.


\section{Conclusions and future work}
\label{conclusion}
\textcolor{black}{
We presented a method for building, with relatively few samples, a custom sampling distribution for random sampling planners. We show evidence that sampling from this custom distribution improves the performance of the RRT planner with respect to a uniform distribution or the RRT goal bias in the proposed tasks. To sample from the custom probabilistic distribution function, we use the rejection sampling method. We also observed that considering the correlation between random variables improves the performance. We tested the method inside five virtual environments using the kinematic and dynamic model for a car-like robot. The selected tasks are common tasks for autonomous vehicles in low-speed and high-speed profiles and even for an environment with narrow passages. The results showed that our method reduced the number of vertices in the tree, especially when the state space is high-dimensional and with obstacles. To be specific, the proposed method decreases the number the vertices in the tree and at the same time increases the likelihood of finding a solution. Based on these experiments, we provide evidence that a custom distribution based on the task improves the performance of sampling-based motion planners. In future work, we will explore techniques that use larger data-sets such as deep-learning to model a custom probability density function in order to deal with changes in the workspace.
}

\section*{Acknowledgements}
This article was supported by the Secretar\'ia de Investigaci\'on y Posgrado of the Instituto Polit\'ecnico Nacional (SIP-IPN) under the research grants, 20210268. Gabriel O. Flores Aquino thank the support from CONACYT.  
%
%


\begin{thebibliography}{}
%
%
\bibitem{b1}LaValle S M (2006) \textsl{Planning Algorithms}. Cambridge University Press
\bibitem{b14}Latombe J C (1991) Robot motion planning. Kluwer Academic Publisher 

\bibitem{b29}Qureshi A H, Miao Y, Simeonov A, Yip M C (2021) Motion Planning Networks: Bridging the Gap Between Learning-Based and Classical Motion Planners. in IEEE Transactions on Robotics, vol. 37, no. 1, pp. 48-66. doi: 10.1109/TRO.2020.3006716

\bibitem{b30}Mohanty P K, Dewang H S (2021) A smart path planner for wheeled mobile robots using adaptive particle swarm optimization. J Braz. Soc. Mech. Sci. Eng. 43, 101. https://doi.org/10.1007/s40430-021-02827-7

\bibitem{b27} Korayem M H, Nazemizadeh M, Nohooji H R (2014) Optimal point-to-point motion planning of non-holonomic mobile robots in the presence of multiple obstacles. J Braz. Soc. Mech. Sci. Eng. 36, 221–232. https://doi.org/10.1007/s40430-013-0063-5

\bibitem{b26}González , Pérez J, Milanés V, Nashashibi F (2016) A Review of Motion Planning Techniques for Automated Vehicles. in IEEE Transactions on Intelligent Transportation Systems, vol. 17, no. 4, pp. 1135-1145. doi: 10.1109/TITS.2015.2498841
\bibitem{b24}LaValle S M (1998) Rapidly-Exploring Random Trees: A New Tool for Path Planning \textsl{Computer Science Dept,} Iowa State University

\bibitem{b5}LaValle S M, Kuffner J J (2001) Randomized Kinodynamic Planning. The International Journal of Robotics Research, 20(5), 378-400. doi:10.1177/02783640122067453
  

\bibitem{b4}Kavraki L E, Svestka P, Latombe JC, Overmars M H (1996) Probabilistic roadmaps for path planning in high-dimensional configuration spaces. in IEEE Transactions on Robotics and Automation, vol. 12, no. 4, pp. 566-580. doi: 10.1109/70.508439

\bibitem{b12}Karaman S, Frazzoli E (2011) Sampling-based algorithms for optimal motion planning. The International Journal of Robotics Research, 30(7), 846-894. doi:10.1177/0278364911406761
  

\bibitem{b10}Li Y, Littlefield Z, Bekris K E (2016) Asymptotically optimal sampling-based kinodynamic planning. The International Journal of Robotics Research, 35(5), 528-564. doi:10.1177/0278364915614386
  

\bibitem{b28}Janson L, Schmerling E, Clark A, Pavone M (2015) Fast marching tree: A fast marching sampling-based method for optimal motion planning in many dimensions. The International Journal of Robotics Research, 34(7), 883-921. doi:10.1177/0278364915577958
  
\bibitem{b6}Chiang H-T L, Hsu J, Fiser M, Tapia L, Faust A (2019) RL-RRT: Kinodynamic Motion Planning via Learning Reachability Estimators From RL Policies. in IEEE Robotics and Automation Letters, vol. 4, no. 4, pp. 4298-4305. doi: 10.1109/LRA.2019.2931199
\bibitem{b13}Faust A et al (2018) PRM-RL: Long-range Robotic Navigation Tasks by Combining Reinforcement Learning and Sampling-Based Planning. IEEE International Conference on Robotics and Automation (ICRA) pp. 5113-5120. doi: 10.1109/ICRA.2018.8461096

\bibitem{b7}Ichter B, Harrison J, Pavone M (2018) Learning Sampling Distributions for Robot Motion Planning. IEEE International Conference on Robotics and Automation (ICRA), pp. 7087-7094. doi: 10.1109/ICRA.2018.8460730

\bibitem{b25} Ichter B and Pavone M (2019) Robot Motion Planning in Learned Latent Spaces. in IEEE Robotics and Automation Letters, vol. 4, no. 3, pp. 2407-2414. doi: 10.1109/LRA.2019.2901898

\bibitem{b32}Wang J, Chi W, Li C, Wang C, Meng M (2020) Neural RRT*: Learning-Based Optimal Path Planning. in IEEE Transactions on Automation Science and Engineering, vol. 17, no. 4, pp. 1748-1758. doi: 10.1109/TASE.2020.2976560

\bibitem{b9}Hsu D, Tingting Jiang, Reif J, Zheng Sun (2003) The bridge test for sampling narrow passages with probabilistic roadmap planners. IEEE International Conference on Robotics and Automation (Cat. No.03CH37422), pp. 4420-4426 vol.3. doi: 10.1109/ROBOT.2003.1242285
\bibitem{b34}Lien JM, Thomas S L, Amato N M (2003) A general framework
for sampling on the medial axis of the free space. IEEE International Conference on Robotics and Automation (Cat. No.03CH37422), pp. 4439-4444 vol.3. doi: 10.1109/ROBOT.2003.1242288

\bibitem{b15}Boor V, Overmars M H, van der Stappen A F (1999) The Gaussian sampling strategy for probabilistic roadmap planners. Proceedings IEEE International Conference on Robotics and Automation (Cat. No.99CH36288C), 1999, pp. 1018-1023 vol.2, doi: 10.1109/ROBOT.1999.772447

\bibitem{b8}Lin Y, (2006) The Gaussian PRM Sampling for Dynamic Configuration Spaces 9th International Conference on Control, Automation, Robotics and Vision, pp. 1-5. doi: 10.1109/ICARCV.2006.345422

\bibitem{b17}  van den Berg J P, Overmars M H (2005) Using Workspace Information as a Guide to Non-uniform Sampling in Probabilistic Roadmap Planners. The International Journal of Robotics Research, 24(12), 1055–1071. https://doi.org/10.1177/0278364905060132

\bibitem{b18}Kurniawati H, Hsu D (2004) Workspace importance sampling for probabilistic roadmap planning. IEEE/RSJ International Conference on Intelligent Robots and Systems (IROS) (IEEE Cat. No.04CH37566), pp. 1618-1623 vol.2. doi: 10.1109/IROS.2004.1389627

\bibitem{b19}Zucker M, Kuffner J, Bagnell J A (2008) Adaptive workspace biasing for sampling-based planners. IEEE International Conference on Robotics and Automation, pp. 3757-3762. doi: 10.1109/ROBOT.2008.4543787

\bibitem{b22}Gammell J D, Srinivasa S S, Barfoot T D (2015) Batch Informed Trees (BIT*): Sampling-based optimal planning via the heuristically guided search of implicit random geometric graphs.  IEEE International Conference on Robotics and Automation (ICRA) pp. 3067-3074. doi: 10.1109/ICRA.2015.7139620
\bibitem{b16}Burns B, Brock O (2007) Single-Query Motion Planning with Utility-Guided Random Trees. Proceedings IEEE International Conference on Robotics and Automation pp. 3307-3312. doi: 10.1109/ROBOT.2007.363983

\bibitem{b20}Yershova A, Jaillet L, Simeon T, LaValle S M (2005) Dynamic-Domain RRTs: Efficient Exploration by Controlling the Sampling Domain. Proceedings of the 2005 IEEE International Conference on Robotics and Automation, pp. 3856-3861. doi: 10.1109/ROBOT.2005.1570709
\bibitem{b21}Berenson D, Abbeel P, Goldberg K (2012) A robot path planning framework that learns from experience. IEEE International Conference on Robotics and Automation pp. 3671-3678. doi: 10.1109/ICRA.2012.6224742
\bibitem{b31}Chamzas C, Shrivastava A, Kavraki L E (2019) Using Local Experiences for Global Motion Planning. International Conference on Robotics and Automation (ICRA) pp. 8606-8612. doi: 10.1109/ICRA.2019.8794317
\bibitem{b23}Coleman D, Şucan I A, Moll M, Okada K, Correll N (2015) Experience-based planning with sparse roadmap spanners. IEEE International Conference on Robotics and Automation (ICRA) pp. 900-905. doi: 10.1109/ICRA.2015.7139284
\bibitem{b33}Kim B, Wang Z, Kaelbling L P, Lozano-Pérez T (2019) Learning to guide task and motion planning using score-space representation. The International Journal of Robotics Research, 38(7), 793–812. https://doi.org/10.1177/0278364919848837
\bibitem{b3}Corke P (2006) \textsl{Robotics, Vision and Control} , 2da ed. Springer, pp.109--111
\bibitem{b2}Bishop C (2006) Pattern Recognition and Machine Learning. Springer, pp.528-530
\bibitem{b11}LaValle S M (2002) Motion Strategy Library. http://msl.cs.uiuc.edu/msl/. Acceded 2019

\end{thebibliography}



\end{document}